\pdfoutput=1

\documentclass[11pt]{article}

\usepackage[]{ACL2023}

\usepackage{times}
\usepackage{latexsym}

\usepackage[T1]{fontenc}

\usepackage[utf8]{inputenc}
\usepackage{times,latexsym}
\usepackage{hyperref}
\usepackage{microtype}
\usepackage{amsmath,amssymb}
\usepackage{epsfig,graphicx,subfigure,caption}
\usepackage{algpseudocode}
\usepackage[normalem]{ulem}
\usepackage{amsmath}
\usepackage[linesnumbered,algoruled,boxed,noend]{algorithm2e}
\usepackage{xcolor}
\usepackage{color, colortbl}
\usepackage{multirow,booktabs, hhline}
\usepackage{amsmath, bm}
\usepackage[linesnumbered,algoruled,boxed,noend]{algorithm2e}
\usepackage{wrapfig}
\usepackage{listings}
\usepackage{url}
\usepackage[T1]{fontenc}

\usepackage{microtype}

\usepackage{inconsolata}

\title{Making Large Language Models Better Data Creators}

\author{
    Dong-Ho Lee\textsuperscript{1}\thanks{~~{Work done during Microsoft Research Internship.}},~
    Jay Pujara\textsuperscript{1},~
    Mohit Sewak\textsuperscript{2},~
    Ryen W. White\textsuperscript{2},~
    Sujay Kumar Jauhar\textsuperscript{2}~\\
    \\
    \textsuperscript{1}Information Sciences Institute, University of Southern California \\
    \textsuperscript{2}Microsoft Research \\
    {\small \texttt{\{dongho.lee\}@usc.edu}, \texttt{\{jpujara\}@isi.edu}, \texttt{\{mohit.sewak,ryenw,sjauhar\}@microsoft.com}}\\
}

\begin{document}
\maketitle
\begin{abstract}

Although large language models (LLMs) have advanced the state-of-the-art in NLP significantly, deploying them for downstream applications is still challenging due to cost, responsiveness, control, or concerns around privacy and security.
As such, trainable models are still the preferred option in some cases.
However, these models still require human-labeled data for optimal performance, which is expensive and time-consuming to obtain.
In order to address this issue, several techniques to reduce human effort involve labeling or generating data using LLMs.
Although these methods are effective for certain applications, in practice they encounter difficulties in real-world scenarios.
Labeling data requires careful data selection, while generating data necessitates task-specific prompt engineering.
In this paper, we propose a unified data creation pipeline that requires only a single formatting example, and which is applicable to a broad range of tasks, including traditionally problematic ones with semantically devoid label spaces.
In our experiments we demonstrate that instruction-following LLMs are highly cost-effective data creators, and that models trained with these data exhibit performance better than those trained with human-labeled data (by up to 17.5\%) on out-of-distribution evaluation, while maintaining comparable performance on in-distribution tasks. These results have important implications for the robustness of NLP systems deployed in the real-world.

\end{abstract}
\section{Introduction}
\begin{figure}[t]
    \centering
    \includegraphics[width=\linewidth]{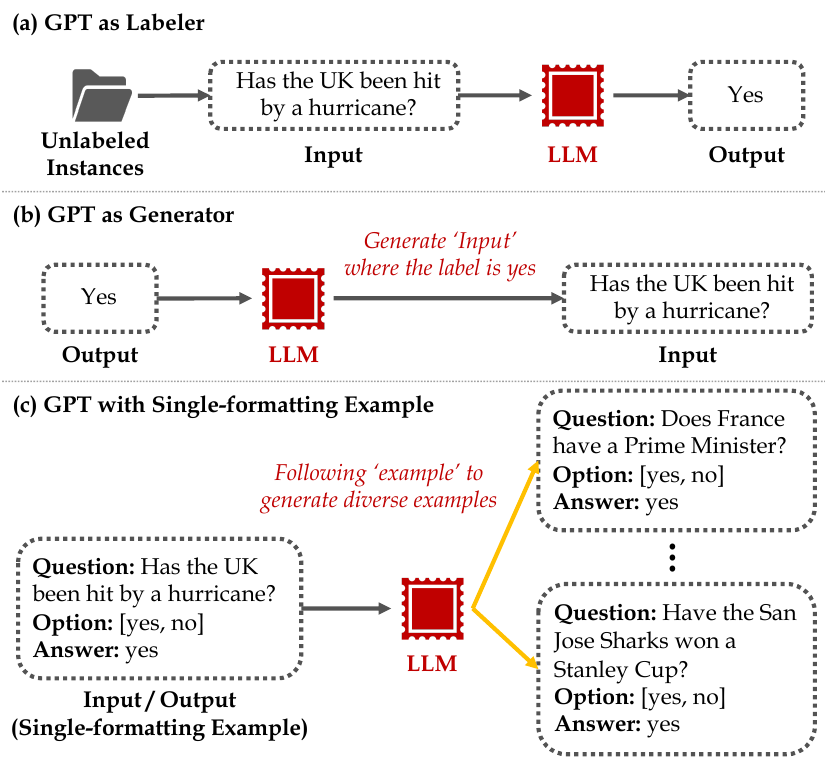}
    \caption{Existing LLM-based data augmentation needs unlabeled examples (labeler) or label-specific prompts (generator), while our framework generates examples for a variety of tasks in a unified way.}
    \label{fig:overview}
    \vspace{-0.5cm}
\end{figure}

Large language models (LLMs) have revolutionized the field of NLP, yielding impressive performance on various conventional natural language understanding (NLU) and generation (NLG) tasks. They are able to do this with only a handful (\textit{i.e.,} few-shot) or sometimes even no training examples (\textit{i.e.,} zero-shot)~\cite{brown2020language, du2022glam, rae2021scaling, thoppilan2022lamda, chowdhery2022palm}.
However, despite their effectiveness, there is a continued demand for the deployment of smaller trainable or tunable models in real-world scenarios due to cost constraints, existing service-level agreement response times, or privacy and security concerns around using black-box APIs.
Unfortunately, application-specific custom models sometimes require large amounts of high-quality human-labeled data, in order to perform well.
Thus, in order to reduce time and cost in the model deployment cycle, recent work has focused on trying to obtain training data by leveraging LLMs as either \textbf{labelers} to annotate unlabeled data~\cite{yoo-etal-2021-gpt3mix-leveraging, wang-etal-2021-want-reduce, lang2022co}, or \textbf{generators} to generate new data samples~\cite{meng2022generating, ye2022zerogen, gao2022zerogen}.


Despite initial successes, constraints for these techniques continue to hinder their applicability in broader real-world settings.
First, in the context of using LLMs as \textbf{labelers}, it is essential to have raw data that closely resembles the distribution of data in the predictive task.
Most previous research has assumed access to a training dataset from which the labels are elided; however, for cold-start problems in the real-world, no such assumptions can be made.
Curating raw data for tasks in specialized domains, such as those in the biomedical or legal fields, can be particularly challenging.
Conversely, sampling a large volume of data at random can result in an imbalanced label distribution due to rare events~\cite{markov2022holistic}. 

Meanwhile, leveraging LLMs as \textbf{generators} requires careful curation of few-shot examples~\cite{hartvigsen-etal-2022-toxigen}, or composition of prompts that highlight the semantic meaning of labels~\cite{wang2021towards, meng2022generating, ye2022zerogen, gao2022zerogen}, such as \emph{positive} v. \emph{negative} in sentiment classification. The latter has been a bottleneck to the broader applicability of LLMs as generators, however, since not all tasks have labels that are semantically meaningful, or are enumerable. Consider, for example the label \emph{yes} v. \emph{no}, which have no meaning when taken without context; or the options of a multiple choice QA (see Figure~\ref{fig:overview}), which are an effectively open-ended label-set that varies from instance to instance. For these kinds of problems LLMs as \textbf{generators} continue to be inadequate.

In this paper, we first present a formal framework for characterizing different approaches for LLM data creation. Specifically, we use graphical models as a way to characterize and unify disparate approaches that include LLMs as either \textbf{labelers} or \textbf{generators} (Section~\ref{sec:formalization}). 
Next, we propose a novel data creation pipeline that only requires a single formatting example to generate heterogeneous labeled data for various downstream applications, including those that focus on specialized domains.
In contrast to current methods that require dataset-specific components (\textit{e.g.,} label description, example selection), our pipeline serves as a unified solution that can be applied to a wide range of tasks, including those where the label set is either semantically devoid of meaning, or unenumerable.

Our data creation pipeline leverages an instruction-following LLM as a \textbf{generator} in conjunction with a single formatting example as a simple yet effective way of imposing structured constraints.
Specifically, our approach iteratively conditions the generator on an instruction and a unique formatting example in a JSON format to yield multiple examples that vary in content but are formatted uniformly (Section~\ref{ssec:instruction}$-$\ref{ssec:formatting-example}).
Furthermore, as an efficient means of diversifying the generated data, we propose a ``self-reference'' strategy, which iteratively samples from the pool of newly created examples to seed the prompt for the next round of generation (Section~\ref{ssec:self-reference}). Specifically, we outline 4 distinct instantiations of ``self-reference'' including \textbf{random}, \textbf{contrastive}, \textbf{similar}, and \textbf{tree} sampling for controlled diversification of data.

We evaluate our data creation pipeline on a battery of tests involving three distinct types of tasks, namely multiple-choice question answering (QA), open-book yes/no QA, and closed-book yes/no QA. The datasets for these tasks range across a variety of domains, including specialized ones such as the biomedical domain. Furthermore, for each category of task, we use a minimum of two datasets in order to compare the out-of-distribution (OOD) generalization of models using original data to synthetically generated LLM data.
Our results demonstrate that leveraging LLMs as generators using our formatting-based creation approach is a highly cost-effective way of creating data that can be effectively used to train models for a variety of downstream tasks, including those in specialized domains, and ones where labels are devoid of semantic meaning or vary across the data.
For in-distribution (ID) settings, naturally having access to large amounts of high-quality manually curated and labeled data is still ideal. However, when only a small amount of human-labeled data is available, our approach yields results that are often comparable, and sometimes even better than the original datasets. This highlights the potential role LLMs can play in the model development cycle, especially in resource-poor or specialized domains.
Further, for the OOD settings, models trained on data generated by our pipeline consistently, and by large margins, outperform their counterparts trained on data from human sources.
This robustness and generalizability has important implications for the deployment of real-world systems that deal with data that are variable, chaotic and often very different from curated academic datasets.
We are realeasing our code and prompts to the community to spur future research in the area\footnote{\href{https://github.com/microsoft/llm-data-creation}{https://github.com/microsoft/llm-data-creation}}.

\section{Formalization of LLM-based data creation}
\label{sec:formalization}
In this section, we attempt to draw different data creation strategies using LLMs into a unified framework, and discuss related research using this framework. 

\subsection{LLM-based data creation}
\label{ssec:data-creation}
Assume a large language model $\mathcal{M}$ (\textit{e.g.,} GPT-3) that has been pre-trained to maximize the likelihood of generating each token in a sequence $\mathbf{x}=\left[x_1, x_2, \ldots, x_n\right]$ 
by conditioning on previous tokens. Then, $\mathcal{M}$ is capable of generating new text through recursive sampling of tokens from its output probability distribution.
Given such a model $\mathcal{M}$ and label space $\mathcal{Y}$, the goal of data creation is to induce samples $(\mathbf{x}, \mathbf{y})$ where $\mathbf{y}\in\mathcal{Y}$.
Based on different instantiations of this general framework, other inputs may be included, such as a label descriptive prompt $\mathcal{W}_\mathbf{y}$ for each $\mathbf{y}\in\mathcal{Y}$, in-domain unlabeled example $\mathbf{x}_u\in\mathcal{D}_{U}$, or a small number of example pairs $(\mathbf{x}_l, \mathbf{y}_l)\in\mathcal{D}_{L}$ along with their corresponding explanation $\mathbf{e}_l$.

\subsection{Formal Framework and Related Work}
\label{ssec:related}

\begin{figure}[t]
    \centering
    \includegraphics[width=0.85\linewidth]{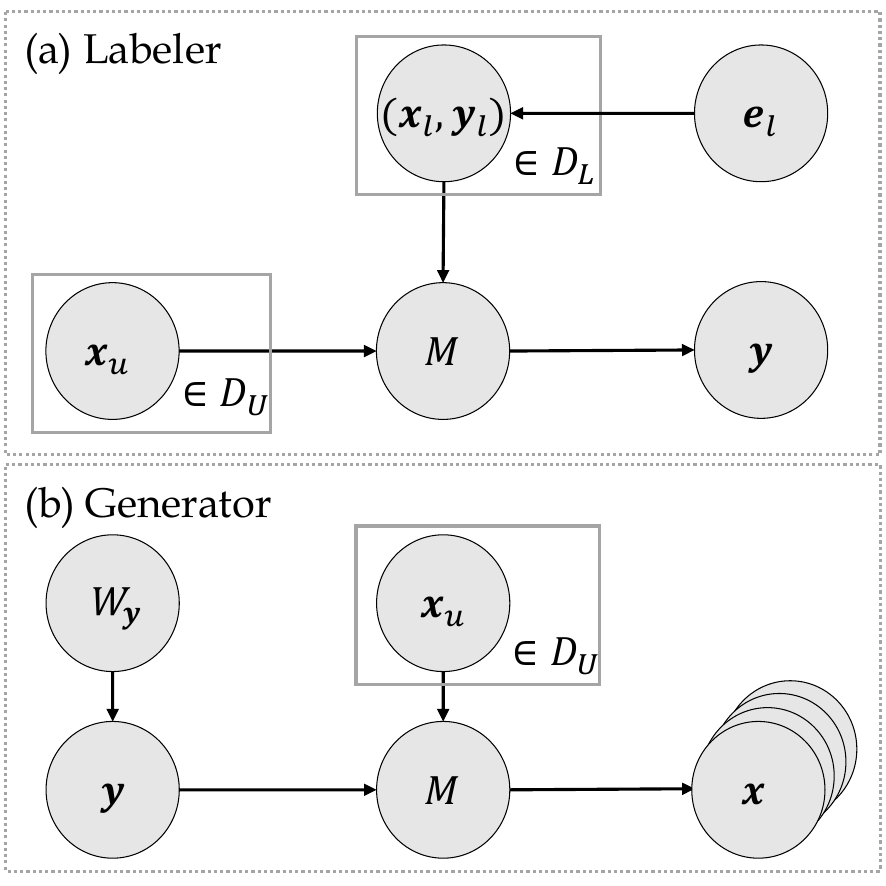}
    \caption{\textbf{Graphical models} for using LLM $\mathcal{M}$ as \textbf{(a) labeler} which outputs label $\mathbf{y}$ for unlabeled data $\mathbf{x}_u\in\mathcal{D}_{U}$ using instructive prompt $\mathcal{W}_\mathbf{y}$ or few-shot examples $(\mathbf{x}_l, \mathbf{y}_l)\in\mathcal{D}_{L}$ with or without explanation $\mathbf{e}_l$,
    and as \textbf{(b) generator} which generates multiple data $\mathbf{x}$ for label $\mathbf{y}$ with label-descriptive prompt $\mathcal{W}$ or using in-domain unlabeled example $\mathbf{x}_u\in\mathcal{D}_{U}$ .
    }
    \label{fig:framework}
    \vspace{-0.5cm}
\end{figure}

Given these basic components, there are two broad strategies for LLM data creation, namely using an LLM as a labeler or as generator. Graphical models for each of these two strategies is presented in Figure~\ref{fig:framework} to summarize the conditional interactions and independence assumptions that describe the relationships between common framework constituents. The rest of this section discusses existing work using the unified language of these graphical models.

\paragraph{Using LLMs as labelers.} 
$\mathcal{M}$ can be used as a labeler for unlabeled data (See Figure~\ref{fig:framework} (a)).
Here, approaches assume that unlabeled data $\mathcal{D}_{U}$ is provided as input and the distribution of $\mathcal{D}_{U}$ is similar to the distribution for the target task.
Labeling can be achieved either by conditioning $\mathcal{M}$ on a few labeled examples $(\mathbf{x}_l, \mathbf{y}_l)\in\mathcal{D}_{L}$~\cite{brown2020language, yoo-etal-2021-gpt3mix-leveraging, wang-etal-2021-want-reduce, lang2022co}, or by leveraging instructive prompts $W$ describing the task without any labeled examples~\cite{brown2020language}.
When providing the model $\mathcal{M}$ with a small number of labeled examples $(\mathbf{x}_l, \mathbf{y}_l)\in\mathcal{D}_{L}$, -- often referred to as the \emph{few-shot} setting (and contrasted with the \emph{zero-shot} setting, where no examples are provided) -- recent studies have shown that curating diverse and representative samples is critical to the ability of the model to label new samples~\cite{liu-etal-2022-makes, rubin-etal-2022-learning, su2022selective}.
This can be challenging, particularly in specialized domains such as, for example, the legal or biomedical domain -- where only a small number of samples may be curated. Our paper proposes a pipeline (Section~\ref{sec:framework} capable of tackling these challenges, and is particularly useful in resource-poor domains (Section~\ref{subsec:perf_comp}).
Furthermore, providing intermediate reasoning steps (\textit{i.e.,} chains-of-thought) as explanations $\mathbf{e}_l$ in prompts enables better labeling in both few-shot~\cite{wei2022chain, zhou2022least, lampinen2022can} and zero-shot setting~\cite{kojima2022large}.

\paragraph{Using LLMs as generators.}
An altogether different data creation approach uses the LLM $\mathcal{M}$ directly as generator (See Figure~\ref{fig:framework} (b)).
While a labeler predicts $\mathbf{y}$ by conditioning on an input $\mathbf{x}$, a generator does the reverse by generating $\mathbf{x}$ given $\mathbf{y}$.
However, like with LLMs as labelers, a small number of relevant samples can be used to condition $\mathcal{M}$ for generating data.
\citet{hartvigsen-etal-2022-toxigen} feeds human-curated examples of the target label (\textit{e.g.,} implicit hate speech) into $\mathcal{M}$ to generate human-like examples for the target label.
In contrast, \citet{wang2021towards} conditions $\mathcal{M}$ on both in-domain unlabeled examples and target label $\mathbf{y}$ to generate domain-related data for $\mathbf{y}$. Meanwhile a number of different efforts~\citep{meng2022generating, ye2022zerogen, gao2022zerogen} condition $\mathcal{M}$ on in-domain unlabeled example and well-formatted descriptive prompts $\mathcal{W}_\mathbf{y}$ for target label $\mathbf{y}$ to generate data.
One important caveat with all these approaches is that the label $\mathbf{y}$, upon which outputs are conditioned, needs to be inherently meaningful in order for instructions to be formulated in a way that prompt the model $\mathcal{M}$ into generating coherent outputs.
For example, when $\mathbf{y}$ is an entailement relationship, a corresponding prompt $\mathcal{W}_\mathbf{y}$ might include ``\textit{$\mathbf{x}_u\in\mathcal{D}_{U}$. In other words...}''; or when $\mathbf{y}$ is the sentiment of a movie review, $\mathcal{W}_\mathbf{y}$ might include ``\textit{The movie review is...}''. 
Contrast this with the scenario where $\mathbf{y}$ is an index or binary response, which has no semantic meaning without context and is therefore difficult to condition on. In this paper, we devise a unified approach to tackling these scenarios (Section~\ref{sec:framework}), yielding a method for data creation that is broadly applicable. 

\paragraph{Differences with Instruction-following Data Generation.}
Recent studies use $\mathcal{M}$ to create data for training instruction-following models~\cite{wang-etal-2023-self-instruct, alpaca, xu2023wizardlm, vicuna2023, mukherjee2023orca}.
Such studies leverage $\mathcal{M}$ as a labeler, producing a coherent response $y$ for a given instruction $x$.
The main focus is on covering a wide range of inputs $x$ and their corresponding responses $y$ to encompass the diversity of instructions potentially encountered in real-world interactions with users.
While recent studies have indicated that such models trained with auto-generated instruction-response pairs can yield logical and coherent responses to user instructions, their performance on NLU tasks remains sub-par~\cite{wang2023far}.
Meanwhile, the focus of our work is the generation of data specifically tailored for natural language understanding (NLU) tasks with a focus on accurate responses.

\begin{figure}[t]
    \centering
    \includegraphics[width=0.85\linewidth]{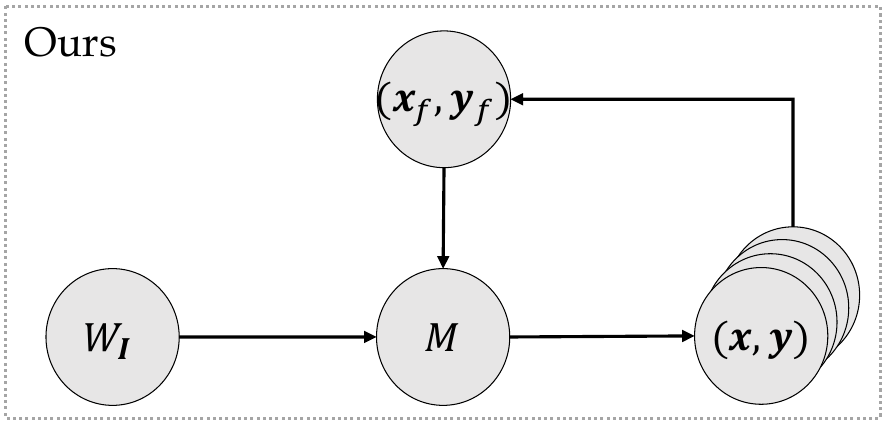}
    \caption{\textbf{Framework Overview} of example-based data creation which outputs multiple complete data $(\mathbf{x}, \mathbf{y})$ starting with an initial seed formatting example $(\mathbf{x}_f, \mathbf{y}_f)$ and the instruction $\mathcal{W}_\mathbf{I}$.
    }
    \label{fig:our_framework}
    \vspace{-0.5cm}
\end{figure}

\section{Example-based Data Creation}
\label{sec:framework}
This paper proposes a unified data creation approach using LLMs, which does not need in-domain unlabeled examples $\mathbf{x}_u \in \mathcal{D}_U$ or data-specific label-descriptive prompts $\mathcal{W}_\mathbf{y}$.
As illustrated in Figure~\ref{fig:our_framework}, our framework iteratively creates data $\mathcal{D}_{G}$ beginning with a single initial formatting example $(\mathbf{x}_f, \mathbf{y}_f)$ and an instruction $\mathcal{W}_\mathbf{I}$ (Section~\ref{ssec:instruction}).
The process begins by converting $(\mathbf{x}_f, \mathbf{y}_f)$ into a structured prompt $\mathcal{W}_f$ (Section~\ref{ssec:formatting-example}-\ref{ssec:formatting-structure}).
After conditioning $\mathcal{M}$ on $[\mathcal{W}_I;\mathcal{W}_f]$ to generate data, we sample an instance from the pool of newly created data to serve as formatting example for the next iteration (Section~\ref{ssec:self-reference}).
We continue to iterate in this manner until the required number of instances $k$ is obtained, after discarding duplicate and ill-formatted data. This is done by caching data and checking newly created candidates against previously generated ones for duplicates, and by using the python \texttt{json.loads()} method for verifying the validity of the json output.
The resulting data creation pipeline is generally suitable for most classification tasks. Although, in this paper we specifically focus on tasks where the label set is potentially open-ended (\textit{e.g.,} multiple-choice QA), or lacks inherent semantic meaning (\textit{e.g.,} binary QA) -- problem spaces that have posed challenges to past work in LLM data creation.

\begin{table}[t]
	\centering
	\scalebox{0.65}{
		\begin{tabular}{l}
            \toprule
            \textbf{Instruction} \\
            \midrule
                \texttt{- You are creating \{$number\_of\_examples$\} examples that} \\
                \texttt{~~follow the format of the example provided,} \\
                \texttt{~~but with a different content.} \\
            \midrule
                \texttt{- The created examples **must** all have different answers.} \\
            \midrule
                \texttt{- The output **must** be in unnumbered JSON format.} 
                \\
            \midrule
                \texttt{- [$fixed\_only$] The created examples **must** have} \\
                \texttt{~~the same options as the provided example.} \\
            \bottomrule
            \end{tabular}
	}
     \caption{\textbf{Instruction} $\mathcal{W}_\mathbf{I}$ used in the paper.}
	\label{tab:prompt}
 \vspace{-0.5cm}
\end{table}

\subsection{Instruction}
\label{ssec:instruction}
The goal of our framework is to have the model $\mathcal{M}$ generate a diverse set of examples in the same format as the input formatting example $(\mathbf{x}_f, \mathbf{y}_f)$.
To ensure format consistency and example diversity, we use the system instruction $\mathcal{W}_\mathbf{I}$ in Table~\ref{tab:prompt}.
We generate data in batches of \{$number\_of\_examples$\}, not only to account for the token generation limits of LLMs, but also to encourage content diversity through subsequent sampling of $\mathcal{W}_f$ (Sec~\ref{ssec:self-reference}).
In this paper, we set \{$number\_of\_examples$\} to 5 and do not vary it.
In order to mitigate label bias, we encourage models to strive for maximal variance in their generated responses, avoiding repetitions in data where the answer is consistently ``yes'', for example.

\subsection{Formatting Example}
\label{ssec:formatting-example}
\begin{figure}[t]
    \centering
    \includegraphics[width=\linewidth]{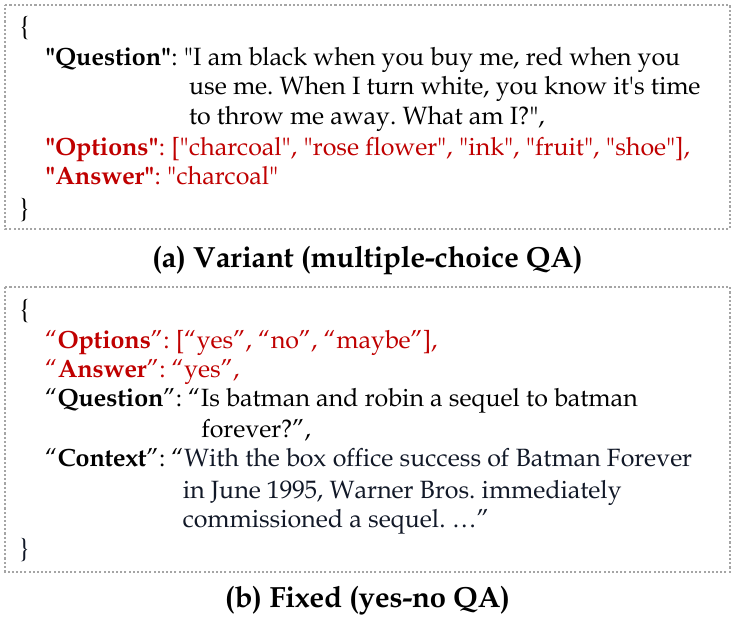}
    \caption{\textbf{Example of formatting example prompt $\mathcal{W}_f$} where ``options'' contain the label space $\mathcal{Y}$ of the task, ``answer'' contains $\mathbf{y}_f$ and ``question'' contains $\mathbf{x}_f$. ``Context'' is optional and depends on the task.}
    \label{fig:format}
\end{figure}

The only assumed input to our example-based data creation pipeline is a single formatting example $(\mathbf{x}_f, \mathbf{y}_f)$ and its corresponding label space $\mathcal{Y}$. This example is formatted as a JSON-structured prompt $\mathcal{W}_f$ as shown in Figure~\ref{fig:format}.
Given the one-shot JSON structured format prompt, it is expected that the model yields a syntactically correct output that conforms to the JSON schema.
While generating a complex structured output like JSON can be challenging, its easy parsing acts as a way to validate outputs at creation time and for effective usage in training downstream models. 

\subsection{Structure of Formatting Example}
\label{ssec:formatting-structure}
Recall that our focus in this paper is on data creation for tasks that are challenging because their output label space is open-ended, or because they lack inherent semantic meaning. We refer to these distinct settings using the shorthand \emph{variable} and \emph{fixed}, and note that the input formatting example $(\mathbf{x}_f, \mathbf{y}_f)$ is different for each of these label space settings.
Specifically, the major difference is the order of presentation of prompt components.

\paragraph{Variable Option.}
The variable option format is structured in a logical sequence beginning with the question $\mathbf{x}_f$, followed by a list of answer candidates $\mathcal{Y}$, and finally the correct answer $\mathbf{y}_f$. 

\paragraph{Fixed Option.}
In contrast, for the variable option, the expected format consists of the answer candidates $\mathcal{Y}$ first, followed by the correct answer $\mathbf{y}_f$, and finally the question $\mathbf{x}_f$.
This inversion of prompt components is added to ensure that the auto-regressive model creates questions with predetermined options
since the model, as a free generator, can produce inconsistent output, resulting in answer options $\mathcal{Y}$ that do not belong to the fixed pre-determined set.

\subsection{Self-Reference}
\label{ssec:self-reference}
Relying on a single formatting example $(\mathbf{x}_f, \mathbf{y}_f)$ as a reference point for all iterations of data creation may limit the ability of the pipeline to yield data that is broad coverage, diverse and balanced. 
To overcome this, we propose ``self-reference'', wherein the formatting example $\mathbf{f}_i=(\mathbf{x}_{f_i}, \mathbf{y}_{f_i})$ for all subsequent generation steps $i > 0$ are sampled from the outputs $(\mathbf{x}_{g_{i-1}}, \mathbf{y}_{g_{i-1}}) \in \mathcal{D}_{G_{i-1}}$ generated at iteration $i-1$.
We experiment with four different sampling strategies.

\paragraph{Random selection.}
During each iteration, a formatting example for the next step is randomly chosen from the output of the current step.

\paragraph{Contrastive selection.}
For each iteration, we select the example that displays the greatest semantic contrast to the preceding formatting example.
In this approach, we use a pre-trained bidirectional-encoder~\cite{reimers-2019-sentence-bert} to generate embeddings for examples, and compute the cosine similarity between $\mathbf{x}_f$ and $\mathbf{x}_{g_{i-1}}$, selecting the instance $\mathbf{x}_{g_{i-1}}$ with the lowest similarity.

\paragraph{Similar selection.}
This sampling approach works analogously to \emph{Contrastive selection}, except that instead of selecting the $\mathbf{x}_{g_{i-1}}$ with the lowest cosine similarity to $\mathbf{x}_f$, we select the one with the highest similarity.


\paragraph{Tree selection.}
Iterative sampling of data may result in significant domain drift from the first seed example to data generated in later steps of the generation pipeline, due to unexpected content variations produced by the model.
To avoid this issue, we use all the generated outputs from one step as formatting examples for subsequent iterations. This approach can be viewed as a breadth-first tree traversal over generated examples, and is in contrast with the other three sampling approaches that use a depth-first exploration strategy. Our hypothesis is that the minimum height of the exploration tree yields samples that are more topically coherent.










\section{Experimental Setup}
In this section we describe the experimental setup that we use to evaluate our single-shot example-based data creation framework.

\subsection{Datasets}

We evaluate on three types of different tasks: multiple-choice question answering (QA), open-book yes/no QA, and closed-book yes/no QA -- as shown in Table~\ref{tab:dataset}.
Multiple-choice QA is used evaluate our data creation pipeline in a variable label space setting, while the other tasks are used for fixed label space settings.
In order to demonstrate the domain generalization capability of our approach, we additionally use a minimum of two datasets for each category of task.
The datasets range broadly in the reasoning abilities they demand from models, requiring them to solve diverse problems such as
filling in the blank in a sentence (PIQA~\cite{Bisk2020}, WinoGrande~\cite{10.1145/3474381}), choosing the most suitable option among multiple choices (CommonsenseQA~\cite{talmor-etal-2019-commonsenseqa}, RiddleSense~\cite{lin-etal-2021-riddlesense}), comprehending a given passage to make a prediction (BoolQ with context~\cite{clark-etal-2019-boolq}, PubMedQA~\cite{jin-etal-2019-pubmedqa}, BioASQ~\cite{tsatsaronis2015overview}), and answering based on inherent knowledge (BoolQ without context~\cite{clark-etal-2019-boolq}, StrategyQA~\cite{geva2021did}, CREAK~\cite{onoe2021creak}).
Details of the various datasets are presented in Appendix~\ref{ssec:dataset_details}.

\subsection{Evaluation Details}
In order to demonstrate the efficacy of our data creation framework, we present a comparison of a downstream model when it is trained on 
(1) the original train dataset denoted by $\mathcal{D}_{L}$; and 
(2) an LLM created dataset, denoted by $\mathcal{D}_{G}$, where a single seed formatting example is randomly selected from $\mathcal{D}_{L}$.
We are unable to conduct a comparative analysis with other LLM-based data generation methods as they do not provide solutions or prompts engineered for the tasks listed in Table~\ref{tab:dataset}.
The base model used in this paper is \texttt{RoBERTa-large}~\cite{liu2019roberta}. 

\subsection{Implementation Details}
Throughout the entire process of data creation, we use \texttt{gpt-3.5-turbo} language model, as of June 2023, with specific settings of $temperature$ and $top_p$ set to 1.
When conducting fine-tuning experiments, we leverage the Adam optimizer~\cite{kingma2014adam} with a maximum sequence length of 256.
In each experiment, we perform a grid search on development data for the optimal learning rate in [3e-4, 1e-4, 5e-5, 2e-5, 1e-5, 5e-6, 3e-6, 1e-6, 5e-7], and batch size in [4, 8, 16]. 
All experiments are conducted on an RTX A5000 with FP32.
\begin{table}[t]
	\centering
	\scalebox{0.63}{
		\begin{tabular}{cccl}
            \toprule
            \textbf{Task} & \textbf{Label Space} & \textbf{Domain} & \textbf{Dataset} \\
            \midrule
            \multirow{2}{*}{multiple-choice QA} & \multirow{2}{*}{Variant (2)} & \multirow{2}{*}{Commonsense} &  PIQA  \\
            & & &  Winogrande  \\
            \midrule
            \multirow{2}{*}{multiple-choice QA} & \multirow{2}{*}{Variant (5)} & \multirow{2}{*}{Commonsense} &  CommonsenseQA  \\
                & & &  RiddleSense  \\
            \midrule
            \multirow{3}{*}{open-book yes/no} & \multirow{3}{*}{Fixed (2)} & Knowledge & BoolQ  (w/ context) \\
            & & Biomedical & PubMedQA \\
            & & Biomedical & BioASQ \\
            \midrule
            \multirow{3}{*}{closed-book yes/no} & \multirow{3}{*}{Fixed (3)} & 
            Knowledge & BoolQ (w/o context) \\
            & & Knowledge & StrategyQA \\
            & & Knowledge & CREAK \\
            \bottomrule
        \end{tabular}
	}
        \caption{\textbf{Datasets used in the paper.} The numbers enclosed in parentheses indicate the number of labels within the label space.}
	\label{tab:dataset}
 \vspace{-0.3cm}
\end{table}


\begin{table*}[t]
    \begin{minipage}{\textwidth}
	\centering
	\small
	\resizebox{\textwidth}{!}{
		\begin{tabular}{rccccccccccccccc}
            \toprule
             & \multicolumn{2}{c}{\textbf{MCQA (2)}} & \multicolumn{2}{c}{\textbf{MCQA (5)}} & \multicolumn{3}{c}{\textbf{Open Yes/No}} & \multicolumn{3}{c}{\textbf{Closed Yes/No}} \\
            \cmidrule(lr){2-3} \cmidrule(lr){4-5} \cmidrule(lr){6-8} \cmidrule(lr){9-11} 
            Trained on $\downarrow$ & PIQA & WinoGrande & CommonsenseQA & RiddleSense & BoolQ & PubMedQA & BioASQ & BoolQ & StrategyQA & CREAK \\
            \midrule
            \# Examples in $\mathcal{D}$ & 14,113 & 160 & 8,500 & 3,510 & 9,427 & 450 & 670 & 9,427 & 2,061 & 10,176 \\
            \midrule
            \midrule
            $\mathcal{D}_{L}$ & 80.95 & 51.41 & 68.17 & 56.48 & 85.62 & 55.20 & 87.14 & 65.68 & 49.56 & 81.19 \\
            \midrule
            $\mathcal{D}_{G}$ (Random) & 66.20 & 51.26 & 42.06 & 37.85 & 68.99 & 59.80 & 80.71 & \underline{52.23} & \underline{53.04} & 67.93 \\
            $\mathcal{D}_{G}$ (Contrastive) & 66.15 & \underline{52.36} & 41.57 & 38.43 & 66.66 & 59.20 & 67.14 & \bf 61.28 & 49.56 & 67.93 \\
            $\mathcal{D}_{G}$ (Similar) & \underline{67.15} & 52.05 & \underline{47.62} & \underline{42.09} & \underline{69.60} & \underline{60.60} & \underline{83.57} & \bf 61.28 & 49.56 & \underline{69.24}  \\
            $\mathcal{D}_{G}$ (Tree) & \bf 68.35 & \bf 52.81 & \bf 48.50 & \bf 42.26 & \bf 69.66 & \bf 61.60 & \bf 85.71 & \bf 61.28 & \bf 56.52 & \bf 72.74  \\
            \midrule
            $(\mathcal{D}_{G}$ - $\mathcal{D}_{L})/\mathcal{D}_{L}$ & $-$18.43\% & $+$2.65\% & $-$40.55\% & $-$33.64\% & $-$22.91\% & $+$10.38\% & $-$1.66\% & $-$7.18\% & $+$12.31\% & $-$11.61\% \\
            \bottomrule
        \end{tabular}
	}
	\end{minipage}
	    \hfill
	\begin{minipage}{\textwidth}
	\centering
	\small
	\end{minipage}
    \vspace{-0.2cm}
	\caption{\textbf{ID Performance (Accuracy)} comparison between models trained on original train dataset $\mathcal{D}_{L}$ (First group) and LLM-created train dataset $\mathcal{D}_{G}$ (Second group). 
    The optimal variant for data-creation in the second group is shown in \textbf{bold}, and the second best is \underline{underlined}.
    The third group of the table presents the percentage difference between the best variant in the second group and the first group.
    }
	\label{tab:main-table}
\end{table*}

\begin{table*}[t]
    \begin{minipage}{\textwidth}
	\centering
	\small
	\resizebox{\textwidth}{!}{
		\begin{tabular}{rccccccccccccccc}
            \toprule
             & \multicolumn{2}{c}{\textbf{MCQA (2)}} & \multicolumn{2}{c}{\textbf{MCQA (5)}} & \multicolumn{4}{c}{\textbf{Open Yes/No}} & \multicolumn{2}{c}{\textbf{Closed Yes/No}} \\
            \cmidrule(lr){2-3} \cmidrule(lr){4-5} \cmidrule(lr){6-9} \cmidrule(lr){10-11} 
            Train $\rightarrow$ & PIQA & WinoGrande & CommonsenseQA & RiddleSense & BoolQ & PubMedQA & BioASQ & PubMedQA & StrategyQA & CREAK \\
            Trained on $\downarrow$ Test $\rightarrow$ & WinoGrande & PIQA & RiddleSense & CommonsenseQA & PubMedQA & BoolQ & PubMedQA & BioASQ & CREAK & StrategyQA \\
            \midrule
            $\mathcal{D}_{L}$ & \bf 52.05 & 44.65 & \underline{41.51} & 40.93 & \underline{62.80} & 58.65 & 67.14 & 56.20 & 49.27 & 48.69 \\
            \midrule
            $\mathcal{D}_{G}$ (Random) & \underline{51.57} & 49.10 & 38.51 & 41.33 & 59.00 & 55.77 & 66.42 &	59.40 &	49.27 &	48.69 \\
            $\mathcal{D}_{G}$ (Contrastive) &  50.31 & 49.50 & 32.94 & 42.35 & 59.00 & 59.87 & 75.00 & 55.20 & 49.27 & 46.95 \\
            $\mathcal{D}_{G}$ (Similar) & 48.42 & \bf 52.25 & \bf 43.42	& \underline{42.62} & \bf 64.60 & \bf 62.50 & \underline{77.85} & \underline{63.00} & 49.27 & \underline{51.30}  \\
            $\mathcal{D}_{G}$ (Tree) & 50.31 & \underline{49.55} & 40.09 & \bf 43.35 & \bf 64.60 & \underline{61.28} &	\bf 81.42 &	\bf 66.00 & \bf 57.72 &	\bf 54.78  \\
            \midrule
            $(\mathcal{D}_{G}$ - $\mathcal{D}_{L})/\mathcal{D}_{L}$ & $-$0.93\% & $+$14.54\% & $+$4.39\% & $+$5.58\% & $+$2.78\% & $+$6.16\% & $+$17.53\% & $+$14.84\% & $+$14.63\% & $+$11.11\% \\
            \bottomrule
        \end{tabular}
	}
	\end{minipage}
	    \hfill
	\begin{minipage}{\textwidth}
	\centering
	\small
	\end{minipage}
    \vspace{-0.2cm}
	\caption{\textbf{OOD Performance (Accuracy)} comparison between models trained on original train dataset $\mathcal{D}_{L}$ (First group) and LLM-created train dataset $\mathcal{D}_{G}$ (Second group).
    The best dataset for each OOD experiment is shown in \textbf{bold}, and the second best is \underline{underlined}.
    The third group of the table presents the percentage difference between the best variant in the second group and the first group.
    }
	\label{tab:transfer-table}
\end{table*}
\section{Experimental Results}
\label{sec:exp_results}
We conduct a comprehensive set of experiments that seek to evaluate the effectiveness of data created by LLMs for focused downstream application modeling. 
Our first evaluation involves examining the performance of models trained on data generated by variants of our single-shot data creation pipeline and comparing them against manually curated training data. We investigate both in-distributed (ID) and out-of-distribution test data settings, each of which are explained in greater detail below.


\subsection{Performance Comparison}
\label{subsec:perf_comp}

\paragraph{ID Performance.}
In the ID setting the test data is drawn from the same distribution as the training data; specifically, a portion of the full human-sourced dataset is held out as a test set to evaluate models trained on either human-labeled or LLM created data.
Table~\ref{tab:main-table} summarizes the performance of models trained on two types of datasets: the original dataset, denoted by $\mathcal{D}_{L}$, and datasets created using the different ``self-reference'' variants of our single-shot pipeline -- the corresponding rows in the table marked as $\mathcal{D}_{G}$. In both settings, and for a given task, the number of samples for each dataset is identical.
We also compute the percentage differential between the two types of datasets, shown in the table as $(\mathcal{D}_{G}$ - $\mathcal{D}_{L})/\mathcal{D}_{L}$. 
These findings confirm that while there is no substitute for large amounts of hand-crafted data, -- demonstrated by drops of up to $-40.55\%$ when using synthetically created data -- LLMs can play an important role when access is only available to very little data, and in specialized domains. This is demonstrated by the similar or often better performance of $\mathcal{D}_{G}$ models on WinoGrande, PubMedQA, BioASQ and StrategyQA.
Meanwhile, a comparison between different ``self-reference'' sampling strategies demonstrates the importance of mitigating domain drift in our single-shot approach, where the sole \emph{true} formatting example is the only anchor point to the data distribution we seek to generate. The \textbf{Tree}-based exploration strategy limits the semantic distance between the seed sample and instances created later in the generation process and therefore yield higher performance on ID data.

\paragraph{OOD Performance.}
While ID data is useful to gain insights of a system in controlled settings, real-world applications must deal with data that is often far more variable and chaotic. Therefore, we compare manually curated data (\textit{i.e.,} original training data) to LLM generated data in an OOD setting. Specifically, since we have used at least two evaluation datasets for each category of downstream application, we use one dataset for training and evaluate on a different dataset. Note that in this setting, while the training data can either be manually curated (i.e. $\mathcal{D}_{L}$), or generated by an LLM (i.e. $\mathcal{D}_{G}$), the test dataset is always manually curated (\textit{i.e.,} original test data).
Table~\ref{tab:transfer-table} presents a comprehensive analysis of the OOD performance of models trained on $\mathcal{D}_{L}$ and $\mathcal{D}_{G}$. 
The results show that models trained on LLM data are consistently and sometimes significantly better at OOD predictive performance than their hand-crafted counterparts. This has important implications for the robustness and generalizability of real-world systems that often deal with inputs that are very different from carefully curated academic datasets. We note that a combination of human and LLM created data may yield even higher gains in OOD performance and leave a comprehensive evaluation of this to future work.
Finally, a comparison of ``self-reference'' strategies on the OOD setting shows that while the \textbf{Tree}-based exploration approach is still a consistently strong one, other sampling approaches are sometimes comparable or better. This is understandable since some degree of controlled noise is helpful, and can be viewed as a regularizer, when trying to generalize to OOD test data.

\subsection{Distribution shift during creation.}
\begin{figure}[t]
    \centering
    \includegraphics[width=\columnwidth]{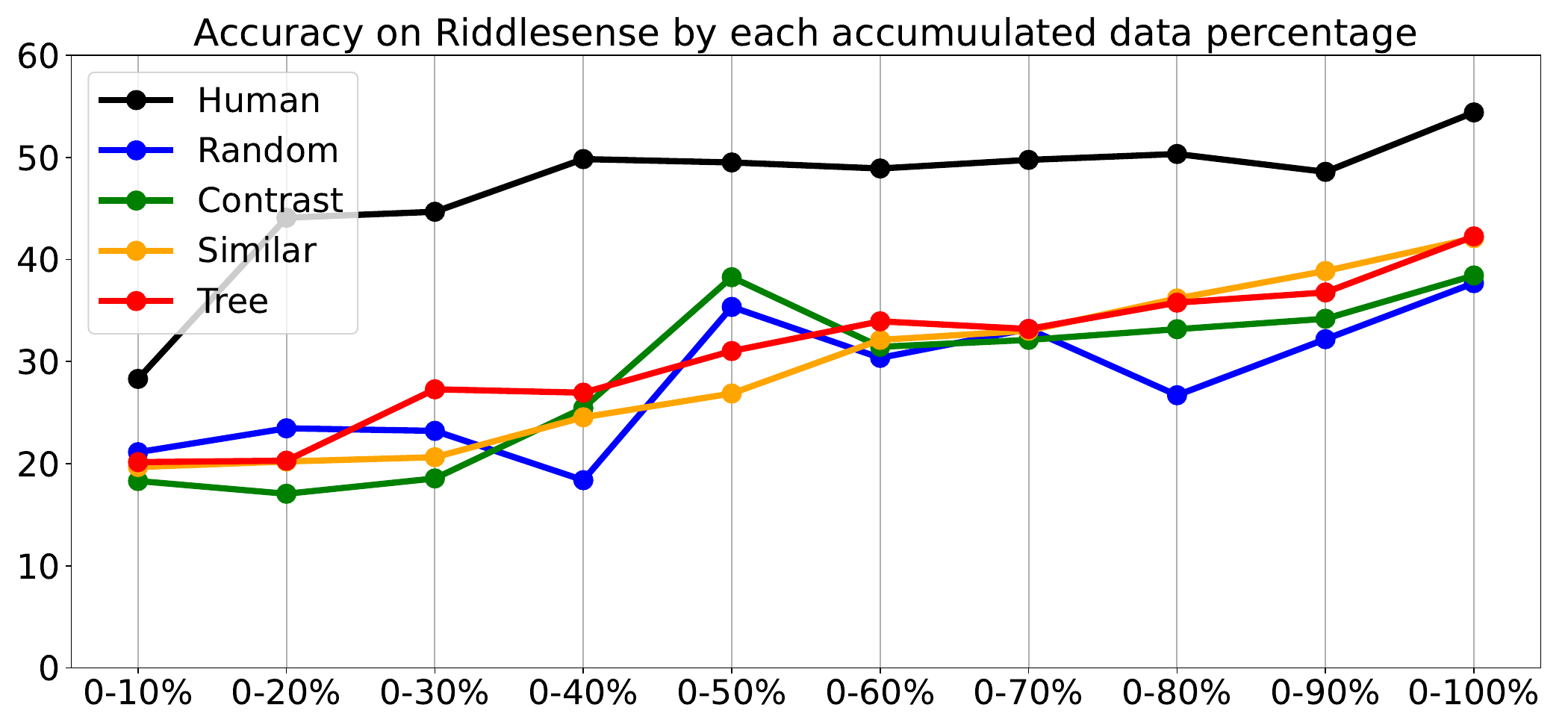}
    \caption{Performance (Accuracy) on RiddleSense using cumulative data splits of the full data.}
    \label{fig:distribution-shift}
    \vspace{-0.3cm}
\end{figure}

One natural question about the different ``self-reference'' strategies is whether the domain drift that they induce from iterative sampling detrimentally impacts samples generated later in the creation process. In other words does inclusion of parts of the dataset generated later lead to performance drops or plateaus?
In order to answer this question we perform an evaluation of cumulative data splits on one of our benchmark datasets (Riddlesense). Specifically we use incremental percentages of training data -- in 10\% blocks -- for all human labeled and synthetically generated datasets, and evaluate the performance of these models on the corresponding test set.
The results of this experiment are shown in Figure~\ref{fig:distribution-shift}.

There are several interesting insights to be gained from these results. Firstly, using human-labeled data leads to much faster convergence; this makes sense since the evaluation is performed on ID test data. \textbf{Random} and \textbf{Contrastive} sampling strategies -- both of which performed less well on our main evaluation -- do exhibit drops in performance with later cumulative splits. Meanwhile, \textbf{Similar} and \textbf{Tree} approaches -- which were consistently better sampling strategies -- demonstrate a steady rise in performance with added data. Jointly these results indicate that judicious selection of examples for subsequent prompts is needed to counter domain drift. Lastly, the final upward trend of all datasets is meaningful because they indicate that models trained on all the datasets do generally benefit from more data. While this additional data is likely difficult and expensive to obtain from human annotators, LLMs can create arbitrarily more data at a fraction of the cost.

\subsection{Data creation cost}

\begin{table}[t]
    \begin{minipage}{0.48\textwidth}
	\centering
	\small
	\resizebox{\textwidth}{!}{
		\begin{tabular}{lc|cccccccccccccc}
            \toprule
            Dataset & \# Train & Random & Diverse & Similar & Tree \\
            \midrule
            PIQA & 14,113 & 3.60 & \bf 2.82 & 3.62 & 3.97 \\
            WinoGrande & 160 & \bf 0.02 & \bf 0.02 & 0.03 & \bf 0.02 \\
            CommonsenseQA & 8,500 & 2.73 & 2.71 & 2.77 & \bf 1.73 \\
            RiddleSense & 3,510 & \bf 0.95 & \bf 0.95 & 1.00 & 1.05\\
            BoolQ & 9,427 & 5.13 & \bf 2.24 & 4.95 & 4.2 \\
            PUbMedQA & 450 & 0.17 & \bf 0.15 & 0.17 & 0.17 \\
            BioASQ & 670 & 0.24 & 0.23 & 0.33 & \bf 0.22 \\
            BoolQ & 9,427 & 3.13 & 4.10 & 3.22 & \bf 3.11 \\
            StrategyQA & 2,061 & 0.66 & 0.70 & 0.81 & \bf 0.66 \\ 
            CREAK & 10,176 & 3.24 & \bf 3.20 & 4.14 & 3.50 \\
            \bottomrule
        \end{tabular}
	}
	\end{minipage}
	    \hfill
	\begin{minipage}{\textwidth}
	\centering
	\small
	\end{minipage}
    \vspace{-0.2cm}
    \caption{\textbf{API Usage Cost (USD)} of data creation strategy. The cost of utilizing API is calculated in USD, based on the current pricing of \texttt{gpt-3.5-turbo} as of June 2023, with a rate of 0.002 USD per 1K tokens. The cheapist strategy is shown in \textbf{bold}.
    }
    \label{tab:cost-table}
\end{table}

Table~\ref{tab:cost-table} presents the expenses incurred by leveraging an instruction-following LLM APIs in our data creation pipeline for each dataset in our evaluation benchmark. 
The results demonstrate that data creation with LLMs is highly cost-effective, and costs well under \$5 USD for every single dataset we created. Factored into this is the cost for data rejected because it was duplicated or ill-formed.
Furthermore, our \textbf{Tree}-based ``self-reference'' strategy -- which was the most performant on quantitative analyses -- was also among the more economical ones. It was the most economical on half the datasets, while the \textbf{Contrastive} strategy incurred the lowest cost on the others.
These expenses are based on the pricing of \texttt{gpt-3.5-turbo} from OpenAI as of June 2023.
\section{Conclusion}
In this paper, we have presented a formal framework for data creation using LLMs and proposed a single-shot formatting example-based data creation pipeline that leverages an instruction-following LLM.
Specifically, we showed how multiple varied examples can be generated from a single seed example in a machine-friendly JSON format by conditioning the LLM on a structured prompt consisting of instructions and a formatted example.
We further expand the diversity of the output by introducing a ``self-reference'' mechanism that selects formatting examples for subsequent iterations of generation from newly created data, and present four different instantiations of the sampling strategy.
While prior efforts at LLM data creation in the literature have avoided domains where the label space is open-ended and varies from instance to instance, or is semantically devoid of inherent meaning, our structured prompts are able to tackle both.
On a battery of evaluations our findings indicate that LLMs can act as highly cost-effective data generators for the development of small trainable or tunable models in downstream applications. For example, a budget of \$5 USD is enough to generate 2M tokens with \texttt{gpt-3.5-turbo}, and depending on the task can yield several thousand data samples.
These models exhibit noteworthy predictive abilities in generalizing to out-of-distribution data, a key desiderata in real-world systems where data can be messy, variable and evolving.
The impact of these findings are meaningful in a number of enterprise scenarios, including for applications that require strict governance on predictive models due to privacy and security concerns, or due to response-time service-level agreements, or indeed for small businesses where human annotation costs, especially from domain experts, can be prohibitively high.

\section{Limitation}
Despite the capability of our pipeline to integrate with a variety of other instruction-following LLMs, our testing is restricted to ChatGPT (\textit{i.e.,} \texttt{gpt-3.5-turbo}) due to performance bottlenecks and time-out issues with LLM APIs in general. While we also experimented with several recent open-source instruction-following models that distill ChatGPT responses, their abilities to generate well-formatted JSON and comprehend instructions were limited. We expect that the integration of our pipeline with other open-source LLMs will be possible in time, as the open-source community attains a performance standard commensurate with commercial products.

An important consideration in our single-shot example-based data creation pipeline is the selection of the initial seed formatting sample. We did not perform an exhaustive analysis to understand the impact of this seed selection on data creation quality, again due to performance bottlenecks with LLM APIs. While we select this seed at random in this paper, it is possible that a more carefully considered approach for crafting or selection of this example may yield better results.

\section{Acknowledgement}
This work was funded in part by the Defense Advanced Research Projects Agency (DARPA) and Army Research Office (ARO) under Contract No. N660011924033, Contract No. W911NF-21-C-0002 and Contract No. HR00112390061, and with support from the Keston Exploratory Research Award.
The views and conclusions contained herein are those of the authors and should not be interpreted as necessarily representing the official policies, either expressed or implied, of DARPA, ARO or the U.S. Government.

\bibliography{anthology,custom}

\begin{thebibliography}{39}
\expandafter\ifx\csname natexlab\endcsname\relax\def\natexlab#1{#1}\fi

\bibitem[{Bisk et~al.(2020)Bisk, Zellers, Bras, Gao, and Choi}]{Bisk2020}
Yonatan Bisk, Rowan Zellers, Ronan~Le Bras, Jianfeng Gao, and Yejin Choi. 2020.
\newblock Piqa: Reasoning about physical commonsense in natural language.
\newblock In \emph{Thirty-Fourth AAAI Conference on Artificial Intelligence}.

\bibitem[{Brown et~al.(2020)Brown, Mann, Ryder, Subbiah, Kaplan, Dhariwal,
  Neelakantan, Shyam, Sastry, Askell et~al.}]{brown2020language}
Tom Brown, Benjamin Mann, Nick Ryder, Melanie Subbiah, Jared~D Kaplan, Prafulla
  Dhariwal, Arvind Neelakantan, Pranav Shyam, Girish Sastry, Amanda Askell,
  et~al. 2020.
\newblock Language models are few-shot learners.
\newblock \emph{Advances in neural information processing systems},
  33:1877--1901.

\bibitem[{Chiang et~al.(2023)Chiang, Li, Lin, Sheng, Wu, Zhang, Zheng, Zhuang,
  Zhuang, Gonzalez, Stoica, and Xing}]{vicuna2023}
Wei-Lin Chiang, Zhuohan Li, Zi~Lin, Ying Sheng, Zhanghao Wu, Hao Zhang, Lianmin
  Zheng, Siyuan Zhuang, Yonghao Zhuang, Joseph~E. Gonzalez, Ion Stoica, and
  Eric~P. Xing. 2023.
\newblock \href {https://lmsys.org/blog/2023-03-30-vicuna/} {Vicuna: An
  open-source chatbot impressing gpt-4 with 90\%* chatgpt quality}.

\bibitem[{Chowdhery et~al.(2022)Chowdhery, Narang, Devlin, Bosma, Mishra,
  Roberts, Barham, Chung, Sutton, Gehrmann et~al.}]{chowdhery2022palm}
Aakanksha Chowdhery, Sharan Narang, Jacob Devlin, Maarten Bosma, Gaurav Mishra,
  Adam Roberts, Paul Barham, Hyung~Won Chung, Charles Sutton, Sebastian
  Gehrmann, et~al. 2022.
\newblock Palm: Scaling language modeling with pathways.
\newblock \emph{arXiv preprint arXiv:2204.02311}.

\bibitem[{Clark et~al.(2019)Clark, Lee, Chang, Kwiatkowski, Collins, and
  Toutanova}]{clark-etal-2019-boolq}
Christopher Clark, Kenton Lee, Ming-Wei Chang, Tom Kwiatkowski, Michael
  Collins, and Kristina Toutanova. 2019.
\newblock \href {https://doi.org/10.18653/v1/N19-1300} {{B}ool{Q}: Exploring
  the surprising difficulty of natural yes/no questions}.
\newblock In \emph{Proceedings of the 2019 Conference of the North {A}merican
  Chapter of the Association for Computational Linguistics: Human Language
  Technologies, Volume 1 (Long and Short Papers)}, pages 2924--2936,
  Minneapolis, Minnesota. Association for Computational Linguistics.

\bibitem[{Du et~al.(2022)Du, Huang, Dai, Tong, Lepikhin, Xu, Krikun, Zhou, Yu,
  Firat et~al.}]{du2022glam}
Nan Du, Yanping Huang, Andrew~M Dai, Simon Tong, Dmitry Lepikhin, Yuanzhong Xu,
  Maxim Krikun, Yanqi Zhou, Adams~Wei Yu, Orhan Firat, et~al. 2022.
\newblock Glam: Efficient scaling of language models with mixture-of-experts.
\newblock In \emph{International Conference on Machine Learning}, pages
  5547--5569. PMLR.

\bibitem[{Gao et~al.(2022)Gao, Pi, Yong, Xu, Ye, Wu, ZHANG, Liang, Li, and
  Kong}]{gao2022zerogen}
Jiahui Gao, Renjie Pi, LIN Yong, Hang Xu, Jiacheng Ye, Zhiyong Wu, WEIZHONG
  ZHANG, Xiaodan Liang, Zhenguo Li, and Lingpeng Kong. 2022.
\newblock Self-guided noise-free data generation for efficient zero-shot
  learning.
\newblock In \emph{The Eleventh International Conference on Learning
  Representations}.

\bibitem[{Geva et~al.(2021)Geva, Khashabi, Segal, Khot, Roth, and
  Berant}]{geva2021did}
Mor Geva, Daniel Khashabi, Elad Segal, Tushar Khot, Dan Roth, and Jonathan
  Berant. 2021.
\newblock Did aristotle use a laptop? a question answering benchmark with
  implicit reasoning strategies.
\newblock \emph{Transactions of the Association for Computational Linguistics},
  9:346--361.

\bibitem[{Hartvigsen et~al.(2022)Hartvigsen, Gabriel, Palangi, Sap, Ray, and
  Kamar}]{hartvigsen-etal-2022-toxigen}
Thomas Hartvigsen, Saadia Gabriel, Hamid Palangi, Maarten Sap, Dipankar Ray,
  and Ece Kamar. 2022.
\newblock \href {https://doi.org/10.18653/v1/2022.acl-long.234} {{T}oxi{G}en: A
  large-scale machine-generated dataset for adversarial and implicit hate
  speech detection}.
\newblock In \emph{Proceedings of the 60th Annual Meeting of the Association
  for Computational Linguistics (Volume 1: Long Papers)}, pages 3309--3326,
  Dublin, Ireland. Association for Computational Linguistics.

\bibitem[{Jin et~al.(2019)Jin, Dhingra, Liu, Cohen, and
  Lu}]{jin-etal-2019-pubmedqa}
Qiao Jin, Bhuwan Dhingra, Zhengping Liu, William Cohen, and Xinghua Lu. 2019.
\newblock \href {https://doi.org/10.18653/v1/D19-1259} {{P}ub{M}ed{QA}: A
  dataset for biomedical research question answering}.
\newblock In \emph{Proceedings of the 2019 Conference on Empirical Methods in
  Natural Language Processing and the 9th International Joint Conference on
  Natural Language Processing (EMNLP-IJCNLP)}, pages 2567--2577, Hong Kong,
  China. Association for Computational Linguistics.

\bibitem[{Kingma and Ba(2014)}]{kingma2014adam}
Diederik~P Kingma and Jimmy Ba. 2014.
\newblock Adam: A method for stochastic optimization.
\newblock \emph{arXiv preprint arXiv:1412.6980}.

\bibitem[{Kojima et~al.(2022)Kojima, Gu, Reid, Matsuo, and
  Iwasawa}]{kojima2022large}
Takeshi Kojima, Shixiang~Shane Gu, Machel Reid, Yutaka Matsuo, and Yusuke
  Iwasawa. 2022.
\newblock Large language models are zero-shot reasoners.
\newblock \emph{arXiv preprint arXiv:2205.11916}.

\bibitem[{Lampinen et~al.(2022)Lampinen, Dasgupta, Chan, Matthewson, Tessler,
  Creswell, McClelland, Wang, and Hill}]{lampinen2022can}
Andrew~K Lampinen, Ishita Dasgupta, Stephanie~CY Chan, Kory Matthewson,
  Michael~Henry Tessler, Antonia Creswell, James~L McClelland, Jane~X Wang, and
  Felix Hill. 2022.
\newblock Can language models learn from explanations in context?
\newblock \emph{arXiv preprint arXiv:2204.02329}.

\bibitem[{Lang et~al.(2022)Lang, Agrawal, Kim, and Sontag}]{lang2022co}
Hunter Lang, Monica~N Agrawal, Yoon Kim, and David Sontag. 2022.
\newblock Co-training improves prompt-based learning for large language models.
\newblock In \emph{International Conference on Machine Learning}, pages
  11985--12003. PMLR.

\bibitem[{Lin et~al.(2021)Lin, Wu, Yang, Lee, and
  Ren}]{lin-etal-2021-riddlesense}
Bill~Yuchen Lin, Ziyi Wu, Yichi Yang, Dong-Ho Lee, and Xiang Ren. 2021.
\newblock \href {https://doi.org/10.18653/v1/2021.findings-acl.131}
  {{R}iddle{S}ense: Reasoning about riddle questions featuring linguistic
  creativity and commonsense knowledge}.
\newblock In \emph{Findings of the Association for Computational Linguistics:
  ACL-IJCNLP 2021}, pages 1504--1515, Online. Association for Computational
  Linguistics.

\bibitem[{Liu et~al.(2022)Liu, Shen, Zhang, Dolan, Carin, and
  Chen}]{liu-etal-2022-makes}
Jiachang Liu, Dinghan Shen, Yizhe Zhang, Bill Dolan, Lawrence Carin, and Weizhu
  Chen. 2022.
\newblock \href {https://doi.org/10.18653/v1/2022.deelio-1.10} {What makes good
  in-context examples for {GPT}-3?}
\newblock In \emph{Proceedings of Deep Learning Inside Out (DeeLIO 2022): The
  3rd Workshop on Knowledge Extraction and Integration for Deep Learning
  Architectures}, pages 100--114, Dublin, Ireland and Online. Association for
  Computational Linguistics.

\bibitem[{Liu et~al.(2019)Liu, Ott, Goyal, Du, Joshi, Chen, Levy, Lewis,
  Zettlemoyer, and Stoyanov}]{liu2019roberta}
Yinhan Liu, Myle Ott, Naman Goyal, Jingfei Du, Mandar Joshi, Danqi Chen, Omer
  Levy, Mike Lewis, Luke Zettlemoyer, and Veselin Stoyanov. 2019.
\newblock Roberta: A robustly optimized bert pretraining approach.
\newblock \emph{arXiv preprint arXiv:1907.11692}.

\bibitem[{Markov et~al.(2022)Markov, Zhang, Agarwal, Eloundou, Lee, Adler,
  Jiang, and Weng}]{markov2022holistic}
Todor Markov, Chong Zhang, Sandhini Agarwal, Tyna Eloundou, Teddy Lee, Steven
  Adler, Angela Jiang, and Lilian Weng. 2022.
\newblock A holistic approach to undesired content detection in the real world.
\newblock \emph{arXiv preprint arXiv:2208.03274}.

\bibitem[{Meng et~al.(2022)Meng, Huang, Zhang, and Han}]{meng2022generating}
Yu~Meng, Jiaxin Huang, Yu~Zhang, and Jiawei Han. 2022.
\newblock Generating training data with language models: Towards zero-shot
  language understanding.
\newblock \emph{arXiv preprint arXiv:2202.04538}.

\bibitem[{Mukherjee et~al.(2023)Mukherjee, Mitra, Jawahar, Agarwal, Palangi,
  and Awadallah}]{mukherjee2023orca}
Subhabrata Mukherjee, Arindam Mitra, Ganesh Jawahar, Sahaj Agarwal, Hamid
  Palangi, and Ahmed Awadallah. 2023.
\newblock Orca: Progressive learning from complex explanation traces of gpt-4.
\newblock \emph{arXiv preprint arXiv:2306.02707}.

\bibitem[{Onoe et~al.(2021)Onoe, Zhang, Choi, and Durrett}]{onoe2021creak}
Yasumasa Onoe, Michael~JQ Zhang, Eunsol Choi, and Greg Durrett. 2021.
\newblock Creak: A dataset for commonsense reasoning over entity knowledge.
\newblock \emph{arXiv preprint arXiv:2109.01653}.

\bibitem[{Rae et~al.(2021)Rae, Borgeaud, Cai, Millican, Hoffmann, Song,
  Aslanides, Henderson, Ring, Young et~al.}]{rae2021scaling}
Jack~W Rae, Sebastian Borgeaud, Trevor Cai, Katie Millican, Jordan Hoffmann,
  Francis Song, John Aslanides, Sarah Henderson, Roman Ring, Susannah Young,
  et~al. 2021.
\newblock Scaling language models: Methods, analysis \& insights from training
  gopher.
\newblock \emph{arXiv preprint arXiv:2112.11446}.

\bibitem[{Reimers and Gurevych(2019)}]{reimers-2019-sentence-bert}
Nils Reimers and Iryna Gurevych. 2019.
\newblock \href {https://arxiv.org/abs/1908.10084} {Sentence-bert: Sentence
  embeddings using siamese bert-networks}.
\newblock In \emph{Proceedings of the 2019 Conference on Empirical Methods in
  Natural Language Processing}. Association for Computational Linguistics.

\bibitem[{Rubin et~al.(2022)Rubin, Herzig, and
  Berant}]{rubin-etal-2022-learning}
Ohad Rubin, Jonathan Herzig, and Jonathan Berant. 2022.
\newblock \href {https://doi.org/10.18653/v1/2022.naacl-main.191} {Learning to
  retrieve prompts for in-context learning}.
\newblock In \emph{Proceedings of the 2022 Conference of the North American
  Chapter of the Association for Computational Linguistics: Human Language
  Technologies}, pages 2655--2671, Seattle, United States. Association for
  Computational Linguistics.

\bibitem[{Sakaguchi et~al.(2021)Sakaguchi, Bras, Bhagavatula, and
  Choi}]{10.1145/3474381}
Keisuke Sakaguchi, Ronan~Le Bras, Chandra Bhagavatula, and Yejin Choi. 2021.
\newblock \href {https://doi.org/10.1145/3474381} {Winogrande: An adversarial
  winograd schema challenge at scale}.
\newblock \emph{Commun. ACM}, 64(9):99–106.

\bibitem[{Su et~al.(2022)Su, Kasai, Wu, Shi, Wang, Xin, Zhang, Ostendorf,
  Zettlemoyer, Smith et~al.}]{su2022selective}
Hongjin Su, Jungo Kasai, Chen~Henry Wu, Weijia Shi, Tianlu Wang, Jiayi Xin, Rui
  Zhang, Mari Ostendorf, Luke Zettlemoyer, Noah~A Smith, et~al. 2022.
\newblock Selective annotation makes language models better few-shot learners.
\newblock \emph{arXiv preprint arXiv:2209.01975}.

\bibitem[{Talmor et~al.(2019)Talmor, Herzig, Lourie, and
  Berant}]{talmor-etal-2019-commonsenseqa}
Alon Talmor, Jonathan Herzig, Nicholas Lourie, and Jonathan Berant. 2019.
\newblock \href {https://doi.org/10.18653/v1/N19-1421} {{C}ommonsense{QA}: A
  question answering challenge targeting commonsense knowledge}.
\newblock In \emph{Proceedings of the 2019 Conference of the North {A}merican
  Chapter of the Association for Computational Linguistics: Human Language
  Technologies, Volume 1 (Long and Short Papers)}, pages 4149--4158,
  Minneapolis, Minnesota. Association for Computational Linguistics.

\bibitem[{Taori et~al.(2023)Taori, Gulrajani, Zhang, Dubois, Li, Guestrin,
  Liang, and Hashimoto}]{alpaca}
Rohan Taori, Ishaan Gulrajani, Tianyi Zhang, Yann Dubois, Xuechen Li, Carlos
  Guestrin, Percy Liang, and Tatsunori~B. Hashimoto. 2023.
\newblock Stanford alpaca: An instruction-following llama model.
\newblock \url{https://github.com/tatsu-lab/stanford_alpaca}.

\bibitem[{Thoppilan et~al.(2022)Thoppilan, De~Freitas, Hall, Shazeer,
  Kulshreshtha, Cheng, Jin, Bos, Baker, Du et~al.}]{thoppilan2022lamda}
Romal Thoppilan, Daniel De~Freitas, Jamie Hall, Noam Shazeer, Apoorv
  Kulshreshtha, Heng-Tze Cheng, Alicia Jin, Taylor Bos, Leslie Baker, Yu~Du,
  et~al. 2022.
\newblock Lamda: Language models for dialog applications.
\newblock \emph{arXiv preprint arXiv:2201.08239}.

\bibitem[{Tsatsaronis et~al.(2015)Tsatsaronis, Balikas, Malakasiotis, Partalas,
  Zschunke, Alvers, Weissenborn, Krithara, Petridis, Polychronopoulos
  et~al.}]{tsatsaronis2015overview}
George Tsatsaronis, Georgios Balikas, Prodromos Malakasiotis, Ioannis Partalas,
  Matthias Zschunke, Michael~R Alvers, Dirk Weissenborn, Anastasia Krithara,
  Sergios Petridis, Dimitris Polychronopoulos, et~al. 2015.
\newblock An overview of the bioasq large-scale biomedical semantic indexing
  and question answering competition.
\newblock \emph{BMC bioinformatics}, 16(1):1--28.

\bibitem[{Wang et~al.(2021{\natexlab{a}})Wang, Liu, Xu, Zhu, and
  Zeng}]{wang-etal-2021-want-reduce}
Shuohang Wang, Yang Liu, Yichong Xu, Chenguang Zhu, and Michael Zeng.
  2021{\natexlab{a}}.
\newblock \href {https://doi.org/10.18653/v1/2021.findings-emnlp.354} {Want to
  reduce labeling cost? {GPT}-3 can help}.
\newblock In \emph{Findings of the Association for Computational Linguistics:
  EMNLP 2021}, pages 4195--4205, Punta Cana, Dominican Republic. Association
  for Computational Linguistics.

\bibitem[{Wang et~al.(2023{\natexlab{a}})Wang, Ivison, Dasigi, Hessel, Khot,
  Chandu, Wadden, MacMillan, Smith, Beltagy et~al.}]{wang2023far}
Yizhong Wang, Hamish Ivison, Pradeep Dasigi, Jack Hessel, Tushar Khot,
  Khyathi~Raghavi Chandu, David Wadden, Kelsey MacMillan, Noah~A Smith,
  Iz~Beltagy, et~al. 2023{\natexlab{a}}.
\newblock How far can camels go? exploring the state of instruction tuning on
  open resources.
\newblock \emph{arXiv preprint arXiv:2306.04751}.

\bibitem[{Wang et~al.(2023{\natexlab{b}})Wang, Kordi, Mishra, Liu, Smith,
  Khashabi, and Hajishirzi}]{wang-etal-2023-self-instruct}
Yizhong Wang, Yeganeh Kordi, Swaroop Mishra, Alisa Liu, Noah~A. Smith, Daniel
  Khashabi, and Hannaneh Hajishirzi. 2023{\natexlab{b}}.
\newblock \href {https://doi.org/10.18653/v1/2023.acl-long.754} {Self-instruct:
  Aligning language models with self-generated instructions}.
\newblock In \emph{Proceedings of the 61st Annual Meeting of the Association
  for Computational Linguistics (Volume 1: Long Papers)}, pages 13484--13508,
  Toronto, Canada. Association for Computational Linguistics.

\bibitem[{Wang et~al.(2021{\natexlab{b}})Wang, Yu, Firat, and
  Cao}]{wang2021towards}
Zirui Wang, Adams~Wei Yu, Orhan Firat, and Yuan Cao. 2021{\natexlab{b}}.
\newblock Towards zero-label language learning.
\newblock \emph{arXiv preprint arXiv:2109.09193}.

\bibitem[{Wei et~al.(2022)Wei, Wang, Schuurmans, Bosma, Chi, Le, and
  Zhou}]{wei2022chain}
Jason Wei, Xuezhi Wang, Dale Schuurmans, Maarten Bosma, Ed~Chi, Quoc Le, and
  Denny Zhou. 2022.
\newblock Chain of thought prompting elicits reasoning in large language
  models.
\newblock \emph{arXiv preprint arXiv:2201.11903}.

\bibitem[{Xu et~al.(2023)Xu, Sun, Zheng, Geng, Zhao, Feng, Tao, and
  Jiang}]{xu2023wizardlm}
Can Xu, Qingfeng Sun, Kai Zheng, Xiubo Geng, Pu~Zhao, Jiazhan Feng, Chongyang
  Tao, and Daxin Jiang. 2023.
\newblock Wizardlm: Empowering large language models to follow complex
  instructions.
\newblock \emph{arXiv preprint arXiv:2304.12244}.

\bibitem[{Ye et~al.(2022)Ye, Gao, Li, Xu, Feng, Wu, Yu, and
  Kong}]{ye2022zerogen}
Jiacheng Ye, Jiahui Gao, Qintong Li, Hang Xu, Jiangtao Feng, Zhiyong Wu, Tao
  Yu, and Lingpeng Kong. 2022.
\newblock Zerogen: Efficient zero-shot learning via dataset generation.
\newblock \emph{arXiv preprint arXiv:2202.07922}.

\bibitem[{Yoo et~al.(2021)Yoo, Park, Kang, Lee, and
  Park}]{yoo-etal-2021-gpt3mix-leveraging}
Kang~Min Yoo, Dongju Park, Jaewook Kang, Sang-Woo Lee, and Woomyoung Park.
  2021.
\newblock \href {https://doi.org/10.18653/v1/2021.findings-emnlp.192}
  {{GPT}3{M}ix: Leveraging large-scale language models for text augmentation}.
\newblock In \emph{Findings of the Association for Computational Linguistics:
  EMNLP 2021}, pages 2225--2239, Punta Cana, Dominican Republic. Association
  for Computational Linguistics.

\bibitem[{Zhou et~al.(2022)Zhou, Sch{\"a}rli, Hou, Wei, Scales, Wang,
  Schuurmans, Bousquet, Le, and Chi}]{zhou2022least}
Denny Zhou, Nathanael Sch{\"a}rli, Le~Hou, Jason Wei, Nathan Scales, Xuezhi
  Wang, Dale Schuurmans, Olivier Bousquet, Quoc Le, and Ed~Chi. 2022.
\newblock Least-to-most prompting enables complex reasoning in large language
  models.
\newblock \emph{arXiv preprint arXiv:2205.10625}.

\end{thebibliography}
\bibliographystyle{acl_natbib}
\clearpage

\appendix

\section{Appendix}
\label{sec:appendix}

\subsection{Details of Dataset}
\label{ssec:dataset_details}

\begin{itemize}
    \item \textbf{PIQA}~\citep{Bisk2020} is a binary-choice question answering task, which chooses the most suitable solution for questions related to physical commonsense.
    \item \textbf{WinoGrande}~\citep{10.1145/3474381} is a task that involves selecting the correct binary option to fill in a given sentence that requires commonsense reasoning.
	\item \textbf{CommonsenseQA}~\citep{talmor-etal-2019-commonsenseqa} is a multiple-choice question answering task, which picks the most appropriate answer on general commonsense questions.
    \item \textbf{Riddlesense}~\citep{lin-etal-2021-riddlesense} is a multiple-choice questions answering task, which picks the most appropriate answer on riddle-style questions that need cognitive process.
    \item \textbf{BoolQ}~\citep{clark-etal-2019-boolq} is a question answering task that answering questions with a simple ``yes'' or ``no'' response.
    Questions are naturally occurring queries sourced from the Google search engine.
    In an open-book setting, the model must comprehend the given context in order to provide an answer, whereas in a closed-book setting, the answer must be provided directly without any context.
    \item \textbf{PubmedQA}~\citep{jin-etal-2019-pubmedqa} is a task that involves answering research questions pertaining to the corresponding abstracts of biomedical research papers, and the answers are provided in the form of ``yes'', ``no'', or ``maybe''. In our study, we treat ``maybe'' as ``no'' to ensure consistent output format with other datasets.
    \item \textbf{BioASQ}~\citep{tsatsaronis2015overview} offers a range of question answering tasks, covering various categories such as factoid, list, summary, and yes/no questions based on the content of biomedical research papers that have been reviewed by experts in the field. For the purpose of this study, our focus will be restricted to questions that have binary answers of ``yes'' or ``no''.   
    \item \textbf{StrategyQA}~\citep{geva2021did} is a benchmark for question-answering that specifically targets open-domain questions where the necessary reasoning path is not explicitly stated in the question, and needs to be inferred through a strategic approach. 
    The answers to these questions are either ``yes'' or ``no''.
    \item \textbf{CREAK}~\citep{onoe2021creak} has been specifically formulated for the purpose of commonsense reasoning pertaining to entity knowledge.
    The dataset comprises assertions of entities, for which the answers need to be specified as either True or False.
\end{itemize}

\subsection{Data Statistics}
\begin{table}[h]
    \centering
    \begin{tabular}{lcccccc}
        \toprule
        \textbf{Dataset} & \# Train & \# Valid & \# Test \\
        \midrule
        PIQA & 14,113 & 1,838 & 2,000 \\
        WinoGrande (XS) & 160 & 633 & 634 \\
        CommonsenseQA & 8,500 & 1,221 & 1,241 \\
        Riddlesense & 3,510 & 1,021 & 1,202\\
        BoolQ & 9,27 & 1,35 & 1,365\\
        PubmedQA & 450 & 50 & 500\\
        BioASQ & 670 & 75 & 140\\
        StrategyQA & 2,061 & 114 & 115 \\
        CREAK & 10,176 & 685 & 686 \\
        \midrule
    \end{tabular}
    
    \caption{\textbf{Data statistics.} Each dataset. We use in-house test set which is randomly splitted from the train set for those dataset that do not provide test set.}
    \label{tab:data-statistics}
    \vspace{-0.3cm}
\end{table}

\end{document}